\documentclass{Latex/Classes/tplp/new_tlp}

\usepackage{Latex/Macros/macros_commons}
\usepackage{Latex/Macros/macros_paper}

\usepackage[nonumberlist]{glossaries}

\newglossaryentry{ap} {
 name={ambiguity propagation},
 plural={ambiguity propagation},
 description={Ambiguity propagation},
}
\newglossaryentry{aspicp} {
 name={{ASPIC\(^{+}\)}},
 description={ASPIC\(^{\mbox{+}}\)},
}
\newglossaryentry{bpmn} {
 name={BPMN},
 plural={BPMN},
 description={Business Process Model and Notation},
}
\newglossaryentry{contractlog} {
 name={ContractLog},
 description={ContractLog},
}
\newglossaryentry{clp} {
 name={CLP},
 plural={CLP},
 description={courteous logic programming},
 first={courteous logic programming (CLP)},
 firstplural={courteous logic programming (CLP)},
}
\newglossaryentry{ctd} {
 name={CTD},
 plural={CTD},
 description={Contrary-to-Duty},
 first={contrary{-}to{-}duty (CTD)},
 firstplural={contrary{-}to{-}duty (CTD)},
}
\newglossaryentry{ctxml} {
 name={CTXML},
 plural={CTXML},
 description={Contract Tracking XML},
 first={\glsentrydesc{ctxml} (\glsentrytext{ctxml})},
 firstplural={\glsentrydesc{ctxml} (\glsentrytext{ctxml})},
}
\newglossaryentry{daml-s} {
 name={DAML-S},
 plural={DAML-S},
 description={DARPA Agent Markup Language for Web Services},
}
\newglossaryentry{datalogp} {
 name={\mbox{Datalog$^{+}$}},
 plural={\mbox{Datalog$^{+}$}},
 description={\mbox{Datalog$^{+}$}},
}
\newglossaryentry{datalogpm} {
 name={\mbox{Datalog$^{\pm}$}},
 plural={\mbox{Datalog$^{\pm}$}},
 description={\mbox{Datalog$^{\pm}$}},
}
\newglossaryentry{deft} {
 name={DEFT},
 plural={DEFT},
 description={Defeasible \mbox{Datalog$^{\pm}$} Tool},
 first={\glsentrytext{deft} (\glsentrydesc{deft})},
 firstplural={\glsentrytext{deft} (\glsentrydesc{deft})},
}
\newglossaryentry{delp} {
 name={DeLP},
 plural={DeLP},
 description={Defeasible Logic Programming},
 first={\glsentrytext{delp} (\glsentrydesc{delp})},
 firstplural={\glsentrytext{delp} (\glsentrydesc{delp})},
}
\newglossaryentry{dl} {
 name={DL},
 description={Defeasible Logic},
 descriptionplural={Defeasible Logics},
 first={\glsentrydesc{dl} (\glsentrytext{dl})},
 firstplural={\glsentrydescplural{dl} (\glsentryplural{dl})},
}
\newglossaryentry{dt} {
 name={defeasible theory},
 plural={defeasible theories},
 description={Defeasible Theory},
 descriptionplural={Defeasible Theories},
}
\newglossaryentry{ec} {
 name={EC},
 plural={EC},
 description={Event Calculus},
 first={\glsentrydesc{ec} (\glsentrytext{ec})},
 firstplural={\glsentrydesc{ec} (\glsentrytext{ec})},
}
\newglossaryentry{fcl} {
 name={FCL},
 plural={FCL},
 description={Formal Contract Language},
 first={\glsentrydesc{fcl} (\glsentrytext{fcl})},
 firstplural={\glsentrydesc{fcl} (\glsentrytext{fcl})},
}
\newglossaryentry{jvm} {
 name={JVM},
 description={Java Virtual Machine},
 descriptionplural={Java Virtual Machines},
 first={\glsentrydesc{jvm} (\glsentrytext{jvm})},
 firstplural={\glsentrydescplural{jvm} (\glsentryplural{jvm})},
}
\newglossaryentry{legalruleml} {
 name={LegalRuleML},
 plural={LegalRuleML},
 description={OASIS LegalRuleML:~\url{https://www.oasis-open.org/committees/legalruleml}},
 descriptionplural={OASIS LegalRuleML:~\url{https://www.oasis-open.org/committees/legalruleml}},
}
\newglossaryentry{lkif} {
 name={LKIF},
 plural={LKIF},
 description={Legal Knowledge Interchange Format},
 first={\glsentrydesc{lkif} (\glsentrytext{lkif})},
 firstplural={\glsentrydesc{lkif} (\glsentrytext{lkif})},
}
\newglossaryentry{mdl} {
 name={MDL},
 description={Modal Defeasible Logic},
 first={\glsentrydesc{mdl} (\glsentrytext{mdl})},
 firstplural={\glsentrydescplural{mdl} (\glsentryplural{mdl})},
}
\newglossaryentry{omg} {
 name={OMG},
 plural={OMG},
 description={Object Management Group},
 first={\glsentrydesc{omg} (\glsentrytext{omg})},
 firstplural={\glsentrydescplural{omg} (\glsentryplural{omg})},
}
\newglossaryentry{owl} {
 name={OWL},
 plural={OWL},
 description={Web Ontology Language},
}
\newglossaryentry{owl-s} {
 name={OWL-S},
 plural={OWL-S},
 description={OWL for Services},
 first={\glsentrydesc{owl-s} (\glsentrytext{owl-s})},
 firstplural={\glsentrydesc{owl-s} (\glsentrytext{owl-s})},
}
\newglossaryentry{penelope} {
 name={PENELOPE},
 description={PENELOPE},
}
\newglossaryentry{ruleml} {
 name={RuleML},
 plural={RuleML},
 description={Rule Markup Language},
 descriptionplural={Rule Markup Language},
 first={\glsentrydesc{ruleml} (\glsentrytext{ruleml})},
 firstplural={\glsentrydescplural{ruleml} (\glsentryplural{ruleml})},
}
\newglossaryentry{sbvr} {
 name={SBVR},
 plural={SBVR},
 description={Semantics of Business Vocabulary and Rules},
 first={\glsentrydesc{sbvr} (\glsentrytext{sbvr})},
 firstplural={\glsentrydescplural{sbvr} (\glsentryplural{sbvr})},
}
\newglossaryentry{semantic web} {
 name={semantic web},
 description={Semantic Web},
}
\newglossaryentry{swrl} {
 name={SWRL},
 plural={SWRL},
 description={Semantic Web Rule Language},
 first={\glsentrydesc{swrl} (\glsentrytext{swrl})},
 firstplural={\glsentrydescplural{swrl} (\glsentryplural{swrl})},
}
\newglossaryentry{tcpc} {
 name={TCPC},
 plural={TCPC},
 description={2016 Telecommunication Consumer Protections Code},
 first={\glsentrydesc{tcpc} (\glsentrytext{tcpc})},
 firstplural={\glsentrydesc{tcpc} (\glsentrytext{tcpc})},
}

\makeglossaries

\usepackage{wrapfig}
\usepackage{hyperref}
\usepackage{cleveref}

\usepackage{changes}
\usepackage{soul,color}
\soulregister\cite7
\soulregister\ref7
\soulregister\pageref7
\newtheorem{definition}{Definition} 

\begin{document}

\long\def\comment#1{}

\newcommand{\papertitle}{Enabling Reasoning with \gls{legalruleml}} 
\title[\papertitle]{\papertitle\thanks{A preliminary version of this paper appeared in the \emph{Proceedings of the 10th International Web Rule Symposium (RuleML 2016)}, J. J. Alferes, L. Bertossi, G. Governatori, P. Fodor and J. Hall, Eds, Springer, 2016, pp.241--257.}\nocite{Lam.2016}}

\author[H.-P. Lam and M. Hashmi]
{HO-PUN LAM and MUSTAFA HASHMI \\
Data61, CSIRO, Australia \\
(e-mail: \href{mailto:brian.lam@data61.csiro.au?subject=LegalRuleML}{brian.lam@data61.csiro.au},
\href{mailto:mustafa.hashmi@data61.csiro.au?subject=LegalRuleML}{mustafa.hashmi@data61.csiro.au})
}

\pagerange{\pageref{firstpage}--\pageref{lastpage}}
\volume{\textbf{10} (3):}
\jdate{March 2002}
\setcounter{page}{1}
\pubyear{2002}

\maketitle

\label{firstpage}

\begin{abstract}

In order to automate verification process,
regulatory rules written in natural language need to be translated into a format 
that machines can understand. 
However, 
none of the existing formalisms can fully represent the elements that appear in legal norms.
For instance, most of these formalisms do not provide features to capture the behavior of deontic effects,
which is an important aspect in automated compliance checking.
This paper presents an approach for transforming legal norms represented using \gls{legalruleml}
to a variant of \glsdesc{mdl} (and vice versa)
such that a legal statement represented using \gls{legalruleml} can be transformed into a machine-readable format
that can be understood
and reasoned about depending upon the client's preferences.






\bigskip
\noindent \textbf{Note:} \emph{This article is currently under consideration for publication in Theory and Practice of Logic Programming (TPLP).}
\end{abstract}



\ignore{\linenumbers*}
\section{Introduction}

Generally, 
regulatory rules written in natural languages are required to transform into machine understandable formalisms
before automated verification can take place.
Over the years,
numerous languages/standards,
such as \gls{ruleml} \cite{RuleML.2000},
\gls{lkif}~\cite{ESTRELLAProject.2008}, 
\gls{sbvr}~\cite{OMG.2008}, 
\gls{penelope}~\cite{Goedertier.2006}, 
ConDec language~\cite{Pesic.2006}, \gls{contractlog}~\cite{Paschke.2005}, 
and \gls{owl-s}\footnote{\gls{owl-s}, 
originally called \gls{daml-s}, 
is an \gls{owl}-based ontology framework
which provides a core set of construct for describing the properties
and capabilities of web services in an unambiguous 
and machine interpretable way.
\url{http://www.daml.org/services/owl-s/}}~\cite{Martin.2004},
have been proposed to facilitate this process. 
Each of these languages offer useful functionalities but is not free from shortcomings of~\cite{Gordon.2009}.
For instance,
\gls{ruleml} is an XML-based standard language
that enables users to use different types of rules
(such as derivation rules, facts, queries, integrity constraints, etc) 
to represent different kinds of elements according to their needs.
However, it lacks support for the use of deontic concepts, 
such as obligations, permissions and prohibitions,
making it impossible to handle cases with \gls{ctd} obligations
(or reparational obligations)~\cite{Carmo:02},
which is not uncommon in legal contracts.


\citeN{Grosof.04} proposed to adopt \gls{clp} as the underlying execution model of \gls{ruleml} rule-base 
(for translating the clauses of a contract).
A rather similar work was the \gls{ctxml} language developed by~\cite{Farrell.04},
with a computational model based on \gls{ec}.
However, 
these studies all suffer from the same problem as they do not consider normative effects.

Later, 
\citeN{Governatori.2005} addressed the shortcomings of \citeANP{Grosof.04}'s work
and extended \gls{dl}~\cite{Nute.2001} with standard deontic operators for representing normative effects 
as well as semantic operator to deal with the \gls{ctd} obligations. 
This extended language also provides \gls{ruleml} compliant data schemas for representing deontic elements and provides constructs to resolve some of the shortcomings that have been 
discussed in~\cite{Gordon.2009}.

Following this line of research,
in this paper we focus on transforming the legal norms represented using \gls{legalruleml}~\cite{LegalRuleML.2013}, 
a normative extension of \gls{ruleml}, 
into a variant of \gls{mdl}~\cite{Governatori.2008a}. 
This is due to the fact
that legal statements are usually described in the form of natural language expressions,
which cannot be applied to information system to automatically process it further.
\gls{legalruleml},
in this sense,
provides a means for the legislators, legal practitioners
and business managers to formalize their legal documents into a machine-readable format
such that information in the documents can be integrated, contrasted and reused,
but direct reasoning with the normative rules in the documents is still not possible.
Hence,
our work reported here makes it possible to use an implementation of \gls{mdl} 
as the engine to compute the extensions on the legal norms represented using \gls{legalruleml}
and reason on them.


The remainder of the paper is structured as follows: 
in~\Cref{section:sampleContract} we tersely discuss a sample contract
which will be used to motivate 
and illustrate how to use the logical framework 
and \gls{legalruleml} in this context.
Then in \Cref{section:background} we will outline the logical framework required to represent contracts
and its normative effects.
\Cref{section:legalruleml} discusses core elements of a \gls{legalruleml} document.
The procedures to transform a legal theory represented using \gls{legalruleml} to~\gls{dl} is discussed in \Cref{section:translation}.
Related work is discussed in \Cref{section:relatedwork} followed by some concluding remarks and pointers for future work.


\section{A Sample Contract}
\label{section:sampleContract}
A \emph{contract} is a set of declarative statements jointly agreed
and performed by all parties
that are involved in a particular task.
It is a branch of the law of obligations
which concerns about the rights and duties 
that arise from the agreed statements.

This paper is based on the analysis of the following sample ``Contract of Services'',
adapted from~\cite{Governatori.2005}.


\begin{center}
{\bf Contract of Services}
\end{center}

\noindent This Deed of Agreement is entered into effects between ABC company (to be known as \purchaser) and ISP plus (to be known as \supplier) WHEREAS \purchaser desires to enter into an agreement to purchase from \supplier the application server (to be known as \goods) in this agreement. Both the parties shall enter into an agreement subject to the following terms and conditions:
{\small 
\setlist[enumerate,2]{label*=\arabic*.,ref=\theenumi.\arabic*,leftmargin=20pt}
\begin{enumerate}[series=myseries,label*=\arabic*.,ref=\arabic*]
\item {\bf Definitions and Interpretations}
	\begin{enumerate}
		\item All prices are in Australian currency unless otherwise explicitly stated.
		\item This agreement is governed by the Australian law and both the parties hereby agree to submit to the jurisdictions of the Courts of the Queensland with respect to this agreement.
	\end{enumerate}
\item {\bf Commencement and Completion}
	\begin{enumerate}
		\item The contract enters into effects as Jan 30, 2016.
		\item The completion date is scheduled as Jan 30, 2017.
	\end{enumerate}
\item {\bf Policy on Price}
	\begin{enumerate}
		\item A {\em ``Premium Customer''} is a customer who has spent more than \$10,000 in goods. Premium Customers are entitled a 5\% discount on new orders.
		\item\label{contract:pricePolicy:specialOrder} Goods marked as {\em ``Special Order''} are subject to a 5\% surcharge. Premium customers are exempt from special order surcharge.
		\item The 5\% discount for premium customers does not apply for goods in promotions.
	\end{enumerate}
\item {\bf Purchase Order}
	\begin{enumerate}
		\item\label{contract:po:price} The \purchaser shall follow the \supplier price lists on the supplier's website.
		\item\label{contract:po:present} The \purchaser shall present \supplier with a purchase order for the provision of \goods within 7 days of the commencement date.
	\end{enumerate}
\item {\bf Service Delivery}
	\begin{enumerate}
		\item The \supplier shall on receipt of a purchase order for \goods make them available within 1 working day.
		\item \goods that are damaged during delivery shall be replaced by the \supplier within 3 working days 
			from the notification by the \purchaser.
			Otherwise,
			the \supplier shall refund the \purchaser
			and pay the \purchaser a penalty of \$1000.
		\item If for any reason the conditions stated in clauses~\ref{contract:po:price} 
			or~\ref{contract:po:present} are not met, 
			the \purchaser is entitled to charge the \supplier at the rate of \$100 for per hour the \goods are not delivered.
	\end{enumerate}
\item {\bf Payments}
	\begin{enumerate}
		\item\label{contract:payments:pay} The payment terms shall be in full upon the receipt of invoice. An interest shall be charged at 5 \% on accounts not paid within 7 days of the invoice date. Another 1.5\% interest shall be applicable if not paid within the next 15 days. The prices shall be as stated in the sales order unless otherwise agreed in writing by the \supplier.
		\item Payments are to be sent electronically, and are to be performed under standards and guidelines outlined in PayPal.
	\end{enumerate}
\item\label{contract:disputes} {\bf Disputes}\qquad \dots 
\item\label{contract:termination} {\bf Termination}\qquad  \dots
\end{enumerate}
}

\medskip
\noindent The agreement\footnote{The contents of clauses \ref{contract:disputes} and \ref{contract:termination} of the agreement have been omitted here 
as they are not relevant to the scope of this paper.} covers a range of rule objectives such as roles of the involved parties 
(e.g., \supplier, \purchaser), authority and jurisdiction (Australia, Queensland Courts), deontic conditions associated with roles (permissions, prohibition), and temporal properties to perform required actions. A contract can be viewed as a legal document containing a finite set of articles (where each article contains a set of clauses and subclauses). The above-discussed agreement includes two main types of clauses namely: 
\begin{enumerate*}[label=(\roman*)]
\item {\em constitutive clauses}, which define the basic concepts contained in this agreement; and  
\item {\em normative/prescriptive clauses}, which regulate the actions of \purchaser and \supplier for the performance of contract, and include deontic notions e.g., obligations, permissions, etc. 
\end{enumerate*}

\section{The Logical Framework}
\label{section:background}

\glsfirst{dl}~\cite{Nute.2001} is a rule-based skeptical approach to non-monotonic reasoning.
It is based on a logic programming-like language
and is a simple, efficient~\cite{Maher.2001}
but flexible formalism capable of dealing with many intuitions of non-monotonic reasoning in a natural 
and meaningful way~\cite{Antoniou.2004a}.
In this section,
we sketch the basics of the logical apparatus used in the paper.
Basically, 
we will combine three logical components, namely: 
\begin{enumerate*}[(i)]
\item defeasible logic, 
\item deontic concepts, and
\item a fragment of logic related to normative violations,
	such as \gls{ctd} obligations.
\end{enumerate*}

The primary use of \gls{dl} in the present context is aimed at 
facilitating the representation of different types of statements in \gls{legalruleml}
into different types of rules according to their nature,
and to resolve the conflicts 
that may arise from the clauses of a contract using priorities 
and override predicates.

\subsection{\glsdesc{dl}}
A \emph{\gls{dt}}~\cite{Antoniou.2001} $D$ as a triple $(F,R,\superior)$, 
where 
\begin{enumerate*}[label=(\roman*)]
\item $F$ is a set of \emph{facts} or \emph{indisputable} statements,
\item $R$ is the set of rules, and
\item $\superior$ is an acyclic \emph{superiority relation} on $R$.
\end{enumerate*} 
Given a set \PROP of \emph{propositional atoms}, 
the set $\LIT=\PROP\cup\set{\neg p\,|\,p\in\PROP}$ denotes the set of \emph{literals}. 
If $q$ is a literal, 
then $\non q$ denotes its complement; if $q$ is a positive literal $p$ 
then $\non q$ is $\neg p$, 
and if $q$ is $\neg p$ then $\non q$ is $p$. 

Hence, given \LBL a set of arbitrary labels, every rule in $R$ is of the form:
\[r:~A(r)\generalrule C(r)\]
where:
\begin{itemize}
\item $r\in\LBL$ is the unique identifier of the rule;
\item $A(r)=\seq{\phi}$, the \emph{antecedent} of the rule, 
	is a finite set of literals denoting the premises of the rule,
	and can be omitted if it is \emph{empty};
\item $\generalrule\in\set{\strict,\defeasible,\defeater}$ denotes the \emph{type} of the rule; 
\item $C(r)$ is the \emph{consequent} (or \emph{head}) of the rule, 
	contains a single literal.
\end{itemize}

The intuition behind different arrows is the following.
\gls{dl} support three types of rules namely: 
\emph{strict rules} ($r:A(r)\strict C(r)$), 
\emph{defeasible rules} ($r:A(r)\defeasible C(r)$) 
and \emph{defeaters} ($r:A(r)\defeater C(r)$).
Strict rules, in the classical sense, are the rules that the conclusion follows every time the antecedents hold;
a defeasible rule is allowed to assert its conclusions in case there is no contrary evidence to it. 
Finally, defeaters suggest there is a connection between its premises and its conclusion(s) 
but not strong enough to warrant the conclusion on its own; 
they are used to defeat rules for the opposite conclusion(s).


\gls{dl} is a \emph{skeptical} nonmonotonic formalism meaning 
that it does not support contradictory conclusions. 
Instead, 
it seeks to resolve conflicts. 
In case there is some support for concluding $A$ 
but there is also support for concluding $\neg A$, 
\gls{dl} does not conclude either of them. 
However, if the support for $A$ is stronger than the support of $\neg A$ 
then $A$ is concluded. 
Here, the superiority relation $\superior$ is used to describe the relative strength of rules on $R$.
When $r_1>r_2$, 
then $r_1$ is called \emph{superior} to $r_2$, 
and $r_2$ \emph{inferior} to $r_1$. 
Intuitively, $r_1>r_2$ expresses that $r_1$ overrides $r_2$ 
if both rules are applicable\footnote{Here the notion of a rule is \emph{applicable} means 
that all the antecedents of the rule are provable;
a rule is \emph{discarded} if at least one of its antecedents is refuted;
a rule is \emph{defeated} if there is a (stronger) rule for the complement of the conclusion that is applicable.}.

\gls{dl} differentiates positive conclusions from negative conclusions,
that is,
literals that can be proved
or literals that are refuted.
In addition,
it is able to determine the strength of conclusions, 
i.e., whether something is concluded using only strict rules and facts,
or whether we have a defeasible conclusion\textemdash a conclusion that can be retracted if more evidence is provided.
Accordingly, 
for a literal $q$,
we have the following four types of conclusions, 
called \emph{tagged} literals:
\begin{itemize}
\item $+\Delta q$ meaning that $q$ is definitely provable in $D$ (i.e., using only facts or strict rules);
\item $-\Delta q$ meaning that $q$ is definitely rejected in $D$;
\item $+\partial q$ meaning that $q$ is defeasibly provable in $D$; and
\item $-\partial q$ meaning that $q$ is defeasibly rejected in $D$.
\end{itemize}

Provability is based on the concept of \emph{derivation} (or \emph{proof}) in $D$ satisfying the proof conditions. 
Informally,
strict derivations are obtained by forward chaining of strict rules 
while a defeasible conclusion $q$ can be derived 
if there is a rule whose conclusion is $q$, 
and its (prerequisite) antecedent has either already been proved 
or given in the case at hand (i.e., facts), 
and any stronger rules whose conclusion is $\neg p$ has prerequisite 
that it failed to be derived. 
In other words, 
a conclusion $q$ is defeasibly derivable when: 
\begin{enumerate*}[label=(\roman*)]
\item $q$ is a fact; or
\item there is an applicable strict or defeasible rule for $q$, 
	and either all rules for $\neg p$ are discarded (i.e., inapplicable) 
	or every rule for $\neg p$ is weaker than an applicable rule for $q$.
\end{enumerate*}

To illustrate the inferential mechanism of \gls{dl},
let us assume we have a \gls{dt} containing the following rules:

\medskip
\begin{tabular}{rr@{}l}
$r_1$: & $SpecialOrder(X)$ & $\defeasible\,\neg Discount(X)$ \\
$r_2$: & $PremiumCustomer(X)$ & $\defeasible\, Discount(X)$ \\
$r_3$: & $Promotion(X)$ & $\defeasible\,\neg Discount(X)$ \\
\end{tabular}

\medskip
\noindent where $>=\set{r_3>r_2, r_2>r_1}$.
The theory states
that products in promotion are not discounted,
and so are special orders except
when the order is placed by a premium customer,
who are normally entitled to a discount (see, clause 3.1. of the contract).

In a scenario where a customer would like to buy a product with special order,
then we can conclude that the price has to be calculated with \emph{no} discount
since rule $r_2$ is not applicable.
In case where the order is received from a premium customer 
and the product is not in promotion,
then the customer is entitled to receive a discount,
as rule $r_2$ is now applicable
and stronger than $r_1$;
while $r_3$,
which is stronger than $r_2$,
is not applicable (i.e., the product is not in promotion).

%


The set of conclusions is finite
and can be computed in linear time~\cite{Maher.2001}.
Besides, 
the reasoning engine can be implemented as a chip~\cite{Song.2008}.
Over the years,
various efficient 
and powerful implementations have been developed~\cite{Maher.2001a,Bassiliades.2004a,Lam.2009,Antoniou-2007} to facilitate the theoretical
and applications development of \gls{dl}.
For a full presentation 
and proof conditions of \gls{dl} please refer to~\cite{Antoniou.2001}.

Recently, some studies have attempted to relate 
\gls{dl} with other logical formalisms through its argumentation semantics~\cite{Governatori.2004}.
For instance,
\citeN{Lam.2016a} have compared the \gls{ap} variant of \gls{dl}~\cite{Antoniou.2000} with \gls{aspicp}~\cite{Modgil.2014,Modgil.2013,Prakken.2010}
based on the acceptability of arguments,
and proposed a mapping from \gls{aspicp} to \gls{dl}.

\citeN{Hecham.2017a},
on the other hand,
proposed a hypergraph-based algorithm for reasoning conclusions 
from existential rules in \emph{Defeasible \gls{datalogpm}}~\cite{Martinez.2014,Deagustini.2015}
\textendash\xspace an extension of \gls{datalogpm}~\cite{Cali.2012}
which includes defeasible facts and defeasible rules,
but allows weak negation instead of classical negation (as in \gls{dl}) in the body of the rules.
Their approach has overcome the non-deterministic issues
that may appear during the reasoning process
and has been implemented as a tool called \gls{deft}.


\subsection{\glsfirst{mdl}}
\label{sec:mdl}

Having the basics of \gls{dl} is not sufficient enough.
The most essential part of developing a legal reasoning system is on creating the framework, norms, etc.,
for representing the normative behavior of a contract.

Here,
we follow the line of work by~\cite{Governatori.2008,Governatori.2008a}
and~\cite{Lam.2013} 
and extend \gls{dl} with the support of modalities.
Let $\MOD$ denotes the set of modal operators
and the set of \emph{modal literals} be $\MODLIT=\set{Xl,\neg Xl\,|\,l\in\LIT,X\in\MOD}$\footnote{Notice that,
here,
we do not allow nesting of modal operators.
This is a simplification aimed at keeping the system manageable,
but does not pose severe limitations for our purpose.}.

To enhance the expressiveness of a rule to encode chains of obligations and violations, 
following the ideas of~\cite{Governatori.2006}, 
a sub-structural operator $\otimes$ is introduced to capture an obligation 
and the obligations arising in response to the violation of the obligation. 
Thus, given an expression like $a\otimes b$, 
the intuitive reading is that if $a$ is possible, 
then $a$ is the first choice and $b$ is the second one; 
if $\neg a$ holds, i.e., $a$ is violated, 
then $b$ is the actual choice. 
That is, 
the $\otimes$-operator is used to build chains of preferences,
called \emph{$\otimes$-expression}, such that: 
\begin{enumerate*}[label=(\roman*)]
\item each literal is an $\otimes$-expression;
\item if $A$ is an $\otimes$-expression and $b$ is a (modal) literal, 
	then $A\otimes b$ is an $\otimes$-expression,
	whose properties are given in~\Cref{definition:otimesExpression} below.
\end{enumerate*}

\medskip
\begin{definition}[\cite{Lam.2013}]\label{definition:otimesExpression}
A $\otimes$-expression is a binary operator satisfying the following properties:
\begin{enumerate}[leftmargin=*]
\item $a\otimes (b\otimes c) =(a\otimes b)\otimes c$ (associativity);
\item $\bigotimes_{i=1}^{n} a_i =
  (\bigotimes_{i=1}^{k-1}a_i) \otimes
  (\bigotimes_{i=k+1}^{n\phantom{k}}a_i)$ where exists $j$ such
  that $a_j= a_k$ and $j<k$ (duplication and contraction).
\end{enumerate}
\end{definition}

And the definition of rule in \gls{mdl} becomes the following.

\begin{definition}
A rule is an expression
\[
r: A(r)\generalrule_{\Box} C(r)
\]
\noindent where
\begin{itemize}[leftmargin=*]
\item $r\in\LBL$ is the unique identifier of the rule;
\item $A(r)=\seq{\phi}$ is a final set of (modal) literals denoting the premises of the rule,
	and can be omitted if it is \emph{empty};
\item $\generalrule\in\set{\strict,\defeasible,\defeater}$ denotes the \emph{type} of the rule 
	and $\Box$ is a modal operator;
\item $C(r)$ is the \emph{consequent} (or \emph{head}) of the rule, 
	which can be either a single (modal) literal,
	or an $\otimes$-expression if $\generalrule\,=\,\defeasible$.
\end{itemize}
\end{definition}

The derivation of conclusions in \gls{mdl} is similar to that of \gls{dl}
but is beyond the scope of this paper.
For a full description of the proof conditions
and algorithms to compute the extensions of \gls{mdl} please refer to~\cite{Governatori.2008a,Lam.2013} for details.

Throughout the paper,
we use the following abbreviations on set of rules: 
$R_s$ ($R_d$) denotes the set of {strict} (defeasible) rules,
$R[q]$ denotes the set of rules with consequent $q$.
For a rule $r\in R$,
we use $C(r,i)$ denotes the $i^{th}$ (modal) literal that appears in $C(r)$,
and $R[c_i=q]$ denotes the set of rules with head $\otimes_i^nc_i$ 
and $c_i=q$ for some $i\in\set{1,n}$.
\section{\gls{legalruleml}: the Legal Rule Markup Language}
\label{section:legalruleml}

\gls{legalruleml}~\cite{OASISLRMLTC.2015} is a rule interchange language proposed by OASIS, 
which extends \gls{ruleml} with features specific to the legal domain~\cite{Athan.2015}. 
It aims to bridge the gap between natural language descriptions 
and semantic norms~\cite{AthanBGPPW.13}, 
and can be used to model various laws, rules 
and regulations by translating the compliance requirements into a machine-readable format~\cite{GHM2015}.

\begin{wrapfigure}{r}{.37\linewidth}\centering

{
\begin{tikzpicture}[
  every node/.style={draw,rounded corners=3pt,inner sep=4pt,font=\sffamily\small\itshape},
  lbl/.style={draw=none},
  wrapper/.style={inner sep=4pt},
  componentNode/.style={}, 
]

\node (association) [text width=8.5em] {Association(s)\bigskip\mbox{}};

\node (context) [lbl,anchor=south west,xshift=0pt,yshift=6pt,inner sep=0pt] at (association.north west) {Context} ;
\begin{scope}[on background layer]
\node (contextWrapper) [componentNode,wrapper,fit=(association)(context)] {};
\end{scope}

\draw let 
	\p1=($(contextWrapper.west)-(contextWrapper.east)$),
	\n1={veclen(\x1,\y1)}
  in 
    node (metadata) [componentNode,anchor=south,yshift=3pt,text width=\dimexpr \n1-8pt] at (contextWrapper.north) {Metadata}
    node (statements) [componentNode,anchor=north,yshift=-3pt,text width=\dimexpr \n1-8pt] at (contextWrapper.south) {Statements\bigskip\mbox{}}
  ;

\node (legalruleml) [anchor=south west,draw=none,inner sep=0pt,yshift=6pt] at (metadata.north west) {LegalRuleML Document} ;

\begin{scope}[on background layer]
\node [wrapper,fit=(legalruleml)(metadata)(context)(statements),fill=white,drop shadow] {};
\node (contextWrapper) [componentNode,wrapper,fit=(association)(context)] {};
\end{scope}

\node (outer) [wrapper,fit=(contextWrapper)(legalruleml)(statements)] {} ;
\node [draw=none] at ($(outer.south)+(0,-.8em)$) {} ;
\end{tikzpicture}

}
 
  \caption{\gls{legalruleml} Document Structure}\vspace{-.6em}
  \label{fig:legalrulemldocumentstructure}
\end{wrapfigure}
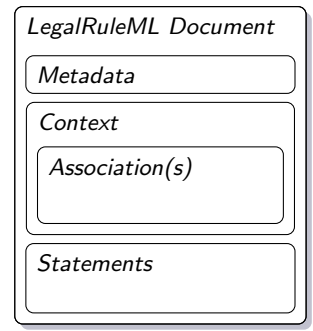
A \gls{legalruleml} document is structured into three main components namely: \emph{metadata}, 
\emph{context}
and \emph{statements},
as depicted in \Cref{fig:legalrulemldocumentstructure}. 
The metadata component contains the legal sources of the norms modelled by the document, 
the temporal information about the legal sources and the document itself, jurisdiction where  
the norms are applicable, and the details concerning the authorities for the legal 
sources and the document. 
The context component, on the other hand, is used to store important relationships 
and corresponding information between metadata and the rules (or fragment of them). 
This is due to the fact that the same rule can be interpreted differently due 
to a variety of parameters such as jurisdiction and temporal parameters that 
can be changed over time. To cater for such situations, 
the description of all characteristics of a particular rule can be stored inside the
context component and will be extracted according to the context as required. 
The statement component contains formal 
representation of the legal norms in the form of rule statements. 



Notice that \gls{legalruleml} supports modelling of defeasibility as within the law.
The intended reading of defeasible rules in \gls{legalruleml} is 
that the conclusion tentatively holds 
when the antecedent of the rule is supported by the evidence/facts of a case.
Then, the conclusion will further be evaluated
if contradictory conclusion(s) with valid arguments has appeared,
or exceptions have been identified.
In addition,
it also provides features to model various effects that follow from applying rules, 
such as obligations, permissions and prohibitions, and can specify preferences among them.






In the next section, 
we will provide the description of how different types of rules 
are represented using \gls{legalruleml}
and describe the way of how they can be transformed into the \gls{dl} framework 
that we have described in the previous section.

\section{Transforming \gls{legalruleml} into \gls{dl}}
\label{section:translation}

A contract written in \gls{legalruleml} is not intended to be executed directly, 
but the business logic can be transformed into a target language of a rule-based system to execute. 
In this section we are going to explore the building blocks of \gls{legalruleml} 
and propose a method to transform legal norms represented in \gls{legalruleml} into a \gls{dl} theory.
Since \gls{legalruleml} is essentially an extension of \gls{ruleml}, 
here we only highlight the differences and identify the additions to faithfully represent legal norms.

\subsection{Premises and Conclusions}
\label{subsec:premisesandconditions}

The first thing we have to consider is the representation of predicates (atoms) to be used in premises 
or conclusions in \gls{legalruleml}. \gls{legalruleml} extends the construct from \gls{ruleml} 
and represents a predicate as an $n$-ary relation, 
and is defined using an element \xmltag[\rulemlprefix]{Atom}\footnote{Elements from \gls{legalruleml}
and elements inherited from \gls{ruleml} will be prefixed with \code{lrml} and \code{ruleml},
respectively.
Information about transforming norms represented using \gls{ruleml} to \gls{dl} can be found in~\cite{Governatori.2005}.
The attributes \code{key}
and \code{keyref} in \gls{legalruleml} correspond to a unique identifier
and reference to an element,
respectively.
Elements inside a \gls{legalruleml} document
can be referenced/link together using these attributes.
However,
since they are not specifically relevant to the discussion here
and thus they are omitted.
}.
Normative effects of an atom, on the other hand,
are captured by embedding the atom inside a deontic element. 
\fussy
The legal concepts such as \emph{obligation} (\xmltag[\lrmlprefix]{Obligation}),
\emph{permission} (\xmltag[\lrmlprefix]{Permission}),
\emph{prohibition} (\xmltag[\lrmlprefix]{Prohibition}), \sloppy
and \emph{right} (\xmltag[\lrmlprefix]{Right})\footnote{Note that the element \emph{right} here is different
from the element ``\emph{right}'' in \gls{ruleml}.
In \gls{legalruleml},
the element ``\emph{right}'' is a deontic specification that gives a permission to a party
and implies
that there is no obligation
or prohibition on the other parties~\cite{OASISLRMLTC.2015};
while the element ``\emph{right}'' in \gls{ruleml} means the right hand side of a rule.}
forms the basic deontic elements in \gls{legalruleml}. 
Further refinements are possible by: 
\begin{enumerate*}[(i)]
\item providing an \code{iri}\footnote{An \code{iri} attribute on a node element in \gls{legalruleml}
  corresponds to an \xmltag[owl]{sameAs} relationship in the abstract syntax.} attribute of a deontic specification,
  or
\item using the \xmltag[\lrmlprefix]{Association} 
and \xmltag[\lrmlprefix]{toTag} elements to link a deontic specification to its meaning 
with the \xmltag[\lrmlprefix]{appliesModality} element\footnote{The \xmltag[\lrmlprefix]{Association} element is used to store the metadata information
that elements in a \gls{legalruleml} theory can associate with;
while the element \xmltag[\lrmlprefix]{toTarget} is used to indicate 
which element(s) the association is going to be applied to.}.

\end{enumerate*}

\begin{lstlisting}[language=XML,
  % ,nolol=true
  % ,caption={Example Right Statement}
  ,label=code:atom,
  frame=none,
  mathescape=true
  ]              
<lrml:Associations>
  <lrml:Association key="asc1">
    <lrml:appliesModality iri="ex:achievementObligation"/>
    <lrml:toTarget keyref="#oblig101"/>
  </lrml:Association>
</lrml:Associations>

<lrml:Obligation key="oblig101">
  <ruleml:Atom key=":atom109">
    <ruleml:Rel iri="pay"/>
    <ruleml:Ind>Purchaser</ruleml:Ind>
    <ruleml:Ind>receivedReciept</ruleml:Ind>
    <ruleml:Ind>Supplier</ruleml:Ind>
  </ruleml:Atom>
</lrml:Obligation>
\end{lstlisting}

\noindent Accordingly, 
\fussy
the above listing represents a modal literal $\literal[\OBL]{pay}[purchaser,receivedReceipt,supplier]$ 
\sloppy
for the clause~\ref{contract:payments:pay} in the contract 
that is true when $purchaser$ has the obligation\footnote{There are several types of obligations based on temporal validity 
and effects they produce e.g., {\em achievement, maintenance} etc., see~\cite{GHM2015} for details.} 
to pay the $supplier$ upon payment receipt\footnote{In this paper, 
we are going to use the modal operator \OBL for obligation, 
\PER for permission, \PRO for prohibition (forbidden).}.


\subsection{Rules and Rulebases}
\label{subsec:rulesandRulebases}

\begin{figure}[t]
\centering

{
\begin{tikzpicture}[level distance=2em,
	level 1/.style={sibling distance=25.em},
	level 2/.style={sibling distance=6.3em},
	every node/.style = {text width=50,align=center,anchor=north},
	l1/.style={yshift=12pt,inner sep=2pt},
	l2/.style={yshift=-3pt},
]
\node (root) {Statements}
	child { node [l1,text width=47] {Normative Statements}
		child { node (cs) [l2] {\begin{varwidth}{10em}\centering Constitutive\\ Statements\end{varwidth}} }
		child { node (ps) [l2] {Prescriptive Statements} }	  	
		}
  	child { node [l1,text width=95] {Violation-Reparation Statements}
	  	child { node (rep) [l2] {Reparation Statements} }
	  	child { node (pen) [l2] {\begin{varwidth}{10em}\centering Penalty\\ Statements\end{varwidth}} }
  		}
  	;

\node (fs) at ($(cs.north)!.4!(pen.north)$) {Factual Statement} ;
\node (os) at ($(cs.north)!.6!(pen.north)$) {Override Statement} ;

\draw (root) -- (fs) ;
\draw (root) -- (os) ;

\end{tikzpicture}
}\\[1.5em]
\caption{Types of Statements in \gls{legalruleml}}
\label{figure:LRMLtypesOfStatements}
\end{figure}
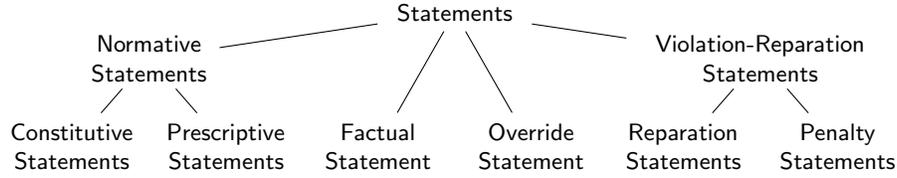

Norms in \gls{legalruleml} are represented as collections of statements, 
and can be classified into four different types according to their nature, 
namely: \emph{norm statements}, \emph{factual statements}, \emph{override statements} 
and \emph{violation-reparation statements}. 
These can be further classified into subtypes, 
as depicted in~\Cref{figure:LRMLtypesOfStatements}. 

\noindent In this section, 
we are going to explore different types of statements 
and describe how they can be transformed into rules in \gls{dl}. 
To facilitate our discussion,
we have the following definition.

\begin{definition}[Compliance and Violation~\cite{OASISLRMLTC.2015}]\label{definition:complianceAndViolation}
\begin{itemize}[leftmargin=*]
\item A \emph{compliance} is an indication that an obligation has been fulfilled 
or a prohibition has not been violated.
\item A \emph{violation} is an indication that an obligation
or prohibition has been violated.
\end{itemize}
\end{definition}


\subsubsection{Norm Statements}
\label{subsec:derivationStatements}
Legal norms, in general, can be classified into \emph{constitutive norms}
(which is used to represent \emph{institutional facts}~\cite{Searle.1997}
and provide definitions of terms and concepts in a
jurisdiction~\cite{OASISLRMLTC.2015},
and \emph{prescriptive norms} (which specify the deontic behavior
and effects of a legal system).
These can be represented as \emph{constitutive statements} (\xmltag[\lrmlprefix]{ConstitutiveStatement})
and \emph{prescriptive statements} in \gls{legalruleml} (\xmltag[\lrmlprefix]{PrescriptiveStatement}),
respectively,
to allow new information to be derived using existing rules.

The following is an example of a prescriptive statement representing the first statement of the clause~\ref{contract:pricePolicy:specialOrder} of the service contract
where \emph{goods} marked with \emph{special order} are subject to a surcharge.

\begin{lstlisting}[language=XML,
  % ,nolol=true
  % ,caption={Example Prescriptive Statement}
  ,label=code:prescriptivestatement,
  frame=none,
  ]
<lrml:PrescriptiveStatement key="r1">
  <ruleml:Rule key=":ruletemplate1">
    <lrml:hasStrength>
      <lrml:DefeasibleStrength key="str1" 
        iri="http://example.org/legalruleml/ontology#defeasible1"/>
    </lrml:hasStrength>
    <ruleml:if>
      <ruleml:And>
        <ruleml:Atom key=":atom2">
          <ruleml:Rel iri=":specialOrder"/>
          <ruleml:Var>X</ruleml:Var>
        </ruleml:Atom>
      </ruleml:And>
    </ruleml:if>
    <ruleml:then>
      <lrml:Obligation>
        <ruleml:Atom key=":atom3">
          <ruleml:Rel iri=":surcharge"/>
          <ruleml:Var>X</ruleml:Var>
        </ruleml:Atom>
      </lrml:Obligation>
    </ruleml:then>
  </ruleml:Rule>
</lrml:PrescriptiveStatement>
\end{lstlisting}

Similar to the \emph{derivation rules} in \gls{ruleml}, 
every constitutive/prescriptive statement has two parts: 
\emph{conditions} (\xmltag[\rulemlprefix]{if}),
which specify the conditions (using a conjunction of formulas and may possibly empty),
and \emph{conclusion} (\xmltag[\rulemlprefix]{then}),
the effects of the rule.
Additionally,
a separate element (\xmltag[\lrmlprefix]{hasStrength}) can be used to specify the strength of the rule.

Both rules can have deontic formulas as their preconditions (body). 
However, 
the difference between the two statements is in the contents of the head, 
where the head of a prescriptive statement is a list of deontic formulas. 
In contrast, 
the head of a constitutive statement cannot be a deontic formula~\cite{OASISLRMLTC.2015}.

In this perspective, a constitutive/prescriptive statement can be transformed into a rule of the form:
\[
 label:~body\generalrule head.
\]

\noindent where $label$ is the key of the statement,
$\generalrule\in\set{\strict,\defeasible,\defeater}$ is the rule type,
$body$ and $head$ are the set of (modal) literals inside the \xmltag[\rulemlprefix]{if} and \xmltag[\rulemlprefix]{then} elements of the statement,
respectively.
Unless otherwise specified,
due to its nature,
the rule modelled using a constitutive statement will be transformed into a strict rule;
while the rule modelled using prescriptive statement will be transformed into a defeasible rule.
Thus, 
the statement above will be transformed to the defeasible rule below\footnote{Note that 
in some variants of \gls{dl}, 
new types of rules can be created for the deontic operator to differentiate between normative 
and definitional rules~\cite{Governatori.2004a}.
For instance, 
the rule $r_1$ above will become: $~\literal{specialOrder}\defeasible_\OBL \literal{surcharge}$ indicating a new type of rule relative to the modal operator $\OBL$. 
However, we do not utilize this approach here as this will limit ourselves 
such that only one type of modality can appear in the head of the rule. 
As it is possible that different logics/semantics can be used to reason on the rules generated using the constitutive 
and prescriptive statements,
using such approach will limit the logic 
that we can use when reasoning the rules. 
For example,
given the skeptical nature of the reasoning,
conclusions of conflicting literals are considered both as not provable
and will ignore the reasons why they were when we use them as premises of further arguments,
which is normally the case for rules generated using constitutive statements.
However,
in some legal settings,
we may want this ambiguity to be propagated along the line of the inference,
which is not uncommon for rules generated using prescriptive statements.
In the first case we speak of ambiguity blocking, 
in the latter case of ambiguity propagation (of \gls{dl}).
As the discussion of this is beyond the scope of this paper,
reader interested in these topics may refer to~\cite{Antoniou.2000,Lam.2011} for details.
}:

\[
r_1:~\literal{specialOrder}[X]\defeasible\literal[\OBL]{surcharge}[X]
\]

\subsubsection{Factual Statements}
\label{subsec:factualstatemetns}
Factual statements,
in essence,
are the expression of facts 
and can be considered as a special case of norm statements without the specification of premises. 
They denote a simple piece of information that is deemed to be true. 
Below is an example of a factual statement in \gls{legalruleml} representing the fact $\literal{premiumCustomer}[JohnDoe]$,
meaning that ``JohnDoe'' is a premium customer.

\begin{lstlisting}[language=XML,
  % ,nolol=true
  % ,caption={Example Right Statement}
  ,label=code:factualStatement,
  frame=none,
  mathescape=true
  ]   
<lrml:FactualStatement key="fact1">
  <lrml:hasTemplate>
    <ruleml:Atom key=":atom11">
      <ruleml:Rel iri=":premiumCustomer"/>
      <ruleml:Ind iri=":JohnDoe"/>
    </ruleml:Atom>
  </lrml:hasTemplate>
</lrml:FactualStatement>
\end{lstlisting}

\subsubsection{Override Statements}
\label{subsec:overridestatements} 
To handle defeasibility, 
\gls{legalruleml} uses \emph{override statements} (\xmltag[\lrmlprefix]{OverrideStatement}) to capture the relative strength of rules
that appear in the legal norms.
The element \xmltag[\lrmlprefix]{Override} defines the relationship of superiority
such that the conclusion of $r2$ overrides the conclusion of $r1$ (where $r1$ and $r2$ are the keys of statements in the legal theory,
as shown below)
if both statements are applicable.


Consider again clause~\ref{contract:pricePolicy:specialOrder} of the contract 
where a {\em premium customer} is exempted from the surcharge for goods marked as {\em `Special Orders'}, 
which can be modelled as the rules below. 

\[
\begin{array}{l@{:~}l}
r_{1} & \literal{specialOrder}[X] \defeasible \literal[\OBL]{surcharge}[X] \\
r_{2} & \literal{specialOrder}[X], \literal{premiumCustomer}[Y] \defeasible \literal[\OBL]{\neg surcharge}[X] \\
\end{array}
\]
\smallskip

\noindent 
In the above example, 
the conclusion of $r_{2}$ takes the precedence over the conclusion of $r_{1}$,
if the order was made from a \emph{premium customer}. 
The following listing illustrates this using an \xmltag[\lrmlprefix]{OverrideStatement} element.

\begin{lstlisting}[language=XML,
  % ,nolol=true
  % ,caption={Example Right Statement}
  ,label=code:overrideStatement,
  frame=none,
  mathescape=true
  ]   
<lrml:OverrideStatement>
  <lrml:Override over="#r2" under="#r1"/>
</lrml:OverrideStatement>
\end{lstlisting}

\noindent
In \gls{dl} terms, this construct defines a superiority relation between  $r_2 > r_1$ 
where $r_1$ and $r_2$ are the rule labels of the rules generated using the statements $r1$ and $r2$ in the legal norms, 
respectively.  

\subsubsection{Violation-Reparation Statements}
Obligations can be violated,
meaning that the content of the obligation has not been achieved.
However, 
a violation may not result in inconsistency
or a termination of interaction
as a penalty can be introduced to compensate the violation~\cite{GHM2015}.

In \gls{legalruleml},
a \emph{Violation-Reparation Statement} is the type of statement concerning what actions are required
when an obligation is violated. 
It provides two basic building blocks to model this,
namely: \emph{penalty statements} (\xmltag[\lrmlprefix]{PenaltyStatement})
and \emph{reparation statements} (\xmltag[\lrmlprefix]{ReparationStatement}),
as shown below.

\noindent\begin{minipage}{.54\textwidth}
\begin{flushright}
\begin{lstlisting}[language=XML,
  basicstyle=\ttfamily\scriptsize,
  % ,nolol=true
  % ,caption={Example Right Statement}
  ,label=code:reparationStatement,
  frame=none,
  mathescape=true,
  ]
<lrml:ReparationStatement key="reps1">
  <lrml:Reparation key="rep1">
    <lrml:appliesPenalty keyref="#pen1"/>
      <lrml:toPrescriptiveStatement 
                 keyref="#ps1"/>
  </lrml:Reparation>
</lrml:ReparationStatement>
\end{lstlisting}
\end{flushright}\end{minipage}
\hfil
\begin{minipage}{.45\textwidth}
\begin{flushleft}
\begin{lstlisting}[language=XML,
  basicstyle=\ttfamily\scriptsize,
  % ,nolol=true
  % ,caption={Example Right Statement}
  ,label=code:penaltyStatement,
  frame=none,
  mathescape=true,
  % xleftmargin=1.2em,
  % framexleftmargin=17pt,
  ]
<lrml:PenaltyStatement key="pen1">
  <lrml:SuborderList>
    list of deontic formulas
  </lrml:SuborderList>
</lrml:PenaltyStatement>
\end{lstlisting}
\end{flushleft}\end{minipage}

\noindent Essentially,
penalty statements model sanctions and/or correction for a violation of a specified rule as outlined in the reparation statement; 
reparation statements bind a penalty statement to the appropriate prescriptive statement 
and apply the penalty when a violation occurs. 
Elements in the \xmltag[\lrmlprefix]{SuborderList} (inside the \xmltag[\lrmlprefix]{PenaltyStatement}) is a list of deontic formulas,
i.e., formula of the form $\literal[Op]A$, where $Op$ is a deontic operator and $A$ is a literal,
such that a formula in the list holds
if all deontic formulas
that precede it in the list have been violated~\cite{OASISLRMLTC.2015}.

To transform these statements into \gls{dl} rules,
we can utilize the $\otimes$-expression that we described in~\Cref{section:background}
by appending the list of modal literals that appear in the penalty statements at the end of original rule.
As an example, 
consider the penalty statement (in clause 6.1 of the contract) 
for not paying invoice within the deadline, 
and assume that the two model literals $\literal[\OBL]{payWith5\% Interest}$ 
and $\literal[\OBL]{payWith6.5\% Interest}$  are transformed from the suborder list inside the penalty statement. 
Then the prescriptive statement $ps1$ will be updated from
\[
{ps1:}~\literal{goods}[X],\literal{invoice}[X] \defeasible \literal[\OBL]{payIn7days}[X]
\]
\noindent to
\[
\begin{array}{p{3.6cm}l}
\multicolumn{2}{l}{
{ps1}: \literal{goods}[X],\literal{invoice}[X]\defeasible \literal[\OBL]{payIn7days}[X]\otimes\literal[\OBL]{payWith5\%Interest}[X]}\\
& \mbox{}\quad\otimes\literal[\OBL]{payWith6.5\%Interest}
\end{array}
\]

\subsection{Other Constructs}
Up to this point, the transformations described have been relatively simple. 
However, 
the transformations cast a wider net than is relevant to the discussion here;
thus,
for our present purpose, 
we limit ourselves to two of the statement/rule-related elements introduced in \gls{legalruleml},
which is not that intuitive.


In legal contract,
there are normative effects,
such as \emph{obligations}, \emph{permissions} and \emph{prohibitions},
that follow from applying rules.
However,
there are situations
where rules are also used to regulate methods for detecting violations
or to determine normative effects triggered by other norm violations,
which are meant to compensate or repair violations~\cite{OASISLRMLTC.2015}.
In this regard
\gls{legalruleml} provides two {\conditionalelements} that can be used to \emph{determine} whether an obligation 
or a prohibition of an object has been fulfilled (\xmltag[\lrmlprefix]{Compliance}) or violated (\xmltag[\lrmlprefix]{Violation}).

Consider the listing below which represents the rule:

\[
ps2: \literal[\PER]{rel1}[X],\literal[\OBL]{rel2}[X]\defeasible\literal[\PRO]{\neg rel3}[X].
\]

\begin{lstlisting}[language=XML,
  % ,nolol=true
  % ,caption={Violation as a condition of a statement}
  ,label=code:ruleViolationBody,
  frame=none,
  mathescape=true
  ]  
<lrml:PrescriptiveStatement key="ps2">
  <ruleml:Rule key=":ruletemplate2">
    <ruleml:if>
      <ruleml:And key=":and1">
        <lrml:Violation keyref="#ps3"/>
        <lrml:Permission>
          <ruleml:Atom key=":atom4">
            <ruleml:Rel iri=":rel1"/>
            <ruleml:Var>X</ruleml:Var>
          </ruleml:Atom>
        </lrml:Permission>
        <lrml:Obligation key="oblig1">
          <ruleml:Atom key=":atom5">
            <ruleml:Rel iri=":rel2"/>
            <ruleml:Var>X</ruleml:Var>
          </ruleml:Atom>
        </lrml:Obligation>
      </ruleml:And>
    </ruleml:if>
    <ruleml:then>
      <lrml:Prohibition key="prohib1">
        <ruleml:Neg key=":neg1">
          <ruleml:Atom key=":atom6">
            <ruleml:Rel iri=":rel3"/>
            <ruleml:Var>X</ruleml:Var>
          </ruleml:Atom>
        </ruleml:Neg>
      </lrml:Prohibition>
    </ruleml:then>
  </ruleml:Rule>
</lrml:PrescriptiveStatement>
\end{lstlisting}

\noindent 
However, 
here we have a violation element appearing in the body as a prerequisite to activate the rule, 
meaning that the referenced element ($ps3$ in this case) has to be violated 
or the rule $ps2$ cannot not be utilised. 
Accordingly, we have two cases:
either \begin{enumerate*}[(i)]
\item the referenced element is a modal literal, or
\item the referenced element is a rule.
\end{enumerate*}

\subsubsection{Case 1: Referenced Element is a literal}

\noindent The former is a simple case. 
If the referenced element is a literal, 
essentially it acts as a precondition to activate the rule. 
It is practically the same as appending the violation (respectively, 
compliance) condition to the body of the rule, 
as shown below.

\[
ps2: \literal[\PER]{rel1}[X],\literal[\OBL]{rel2}[X],\violate{p}\defeasible\literal[\PRO]{\neg rel3}[X].
\]

\noindent where $p$ is the referenced literal, $\violate{p}$ (respectively $\comply{p}$) is a transformation, 
as defined in~\Cref{table:literalTransformationsUnderComplianceAndViolation}, 
that transforms the (modal) literal $p$ into a set of literals 
that needs to be derived in order to satisfy the condition of violation (compliance). 
This is due to the fact 
that, 
basically, 
for a literal $q$,
a situation is violated when we have $\literal[\OBL]{q}$ and $\literal{\neg q}$ (for obligation), 
or $\literal[\PRO]{q}$ and $\literal{q}$ (for forbidden or prohibition);
while a situation is compliance when we have $\literal[\OBL]{q}$ and $\literal{q}$ (for obligation),
or $\literal[\PRO]{q}$ and $\literal{\neg q}$ (for forbidden or prohibition).
For instance, 
if $ps3$ is the modal literal $\literal[\OBL]{q}$, 
then the rule $ps2$ above will be updated as follows

\[
ps2: \literal[\PER]{rel1}[X],\literal[\OBL]{rel2}[X],\convertviolate{\OBL}{q}\defeasible\literal[\PRO]{\neg rel3}[X].
\]

\begin{table}[!t]
\centering
\caption{Requirements to determine whether a literal is compliant  or violated.}\medskip
\begin{tabular}{l@{\qquad}c@{\qquad}c@{\qquad}c}
\toprule
  & \multicolumn{1}{c}{$q$} & \multicolumn{1}{c}{\literal[\OBL]{q}} & \multicolumn{1}{c}{\literal[\PRO]{q}} \\
\midrule
\text{Compliance} & \literal{q} & \convertcomply{\OBL}{q} & \convertcomply{\PRO}{q}  \\
\text{Violation} & \literal{\neg q} & \convertviolate{\OBL}{q} & \convertviolate{\PRO}{q} \\
\bottomrule
\end{tabular}
\label{table:literalTransformationsUnderComplianceAndViolation}
\end{table}

However,
the case is somewhat complex when the referenced \conditionalelement appears at the head of the statement, 
as shown in the listing below.
\begin{lstlisting}[language=XML,
  % ,nolol=true
  % ,caption={Violation as a nested element in the head of a statement}
  ,label=code:ruleViolationHead,
  frame=none,
  mathescape=true,
  escapeinside={(@}{@)}
  ]  
<lrml:PrescriptiveStatement key="ps4">
  <ruleml:Rule key=":ruletemplate3" keyref=":ruletemplate2">
    <ruleml:if>
      $\vdots$
    <ruleml:if>
    <ruleml:then>
      <lrml:SuborderList>
        <lrml:Obligation key="obl1"> (@\label{listing:ruleViolation:literal1:start}@)
          <ruleml:Atom key=":atom26">
            <ruleml:Rel iri=":rel3"/>
            <ruleml:Var>X</ruleml:Var>
          </ruleml:Atom>
        </lrml:Obligation> (@\label{listing:ruleViolation:literal1:end}@)
        <ruleml:And>
          <lrml:Violation keyref="#ps5"/> (@\label{listing:ruleViolationReferenceObject}@)
          <lrml:Obligation key="obl2"> (@\label{listing:ruleViolationHeadOblLiteral:start}@)
            <ruleml:Atom key=":atom27">
              <ruleml:Rel iri=":rel4"/>
              <ruleml:Var>X</ruleml:Var>
            </ruleml:Atom>
          </lrml:Obligation> (@\label{listing:ruleViolationHeadOblLiteral:end}@)
        </ruleml:And>
      </lrml:SuborderList>
    </ruleml:then>
  </ruleml:Rule>
</lrml:PrescriptiveStatement>
\end{lstlisting}

\noindent Here, $\literal[\OBL]{rel4}$ \mbox{(Lines~\ref{listing:ruleViolationHeadOblLiteral:start}-\ref{listing:ruleViolationHeadOblLiteral:end})} 
is derivable only when the modal literal $\literal[\OBL]{rel3}$ 
\mbox{(Lines~\ref{listing:ruleViolation:literal1:start}-\ref{listing:ruleViolation:literal1:end})} 
is defeated and the reference literal $ps5$ \mbox{(\Cref{listing:ruleViolationReferenceObject})} is violated,
which can be considered as a precondition of making $\literal[\OBL]{rel4}$ becomes applicable
and can be represented as the rule below:
\[
ps4: A(ps4)\defeasible\literal[\OBL]{rel3}\otimes(\violate{ps5}\defeasible\literal[\OBL]{rel4})
\]

\noindent 
where $A(ps4)$ is the antecedent of the rule $ps4$.
Notice that, here we have abused the use of notations by nesting a sub-rule into 
the head of the rule. However, such nested structure is \emph{not} supported semantically 
in \gls{dl}. To resolve this issue, we 
have to modify the statement based on its expanded form. 

\begin{definition}[$\otimes$-expansion]\label{definition:ruleExpansion}
Let $D=(F,R,>)$ be a \gls{dt},
and let $\Sigma$ be the language of $D$. 
We define $reduct(D)=(F,R',>')$ where for every rule $r\in R_d$ with a $\otimes$-expression
\opcsseq{\otimes}{c}{1},
appears in its head:
\[
\begin{array}{l}
R'=R \setminus R_d \cup \{
\begin{array}[t]{rl}
r: & \literal{A}[r]\defeasible \literal{c_1} \\[.2em]
r': & \literal{A}[r],\violate{c_1}\defeasible\opcsseq{\otimes}{c}{2}\} \\[.3em]
\end{array} \\
\forall r',s'\in R', r'\superior s' \Leftrightarrow r,s\in R\text{~s.t.~}r'\in reduct(r), 
s'\in reduct(s), r>s.
\end{array}
\]
\end{definition}
\begin{definition}\label{definition:verificationRuleGeneration}
Let $D=(F,R,>)$ be a \gls{dt},
$\Sigma$ be the language of $D$,
and $\forall r\in R_d$,
$C(r)=p$ is a (modal) literal.
We define $\mathcal{T}(D)=(F,R',>)$ where $\forall r\in R_d$:
\[
R'= R\setminus R_d \cup \set{r: A(r),verify(p)\defeasible p}
\]

\noindent where $\verify{p}$ is defined as:
\begin{equation*}
\begin{cases}
  \violate{e} & \text{if a violation element is attached to the element~}p, \\
  \comply{e} & \text{if a compliance element is attached to the element~}p, \\
  \emptyset & \text{otherwise.}
  \end{cases}
\end{equation*}
\indent where $e$ is the literal referenced by the \conditionalelement attributed to $C$.
\end{definition}

\noindent Here, we can first exclude the \conditionalelements in the rule head 
and generate the rule based on $\otimes$-expression.
Then,
we can apply \Cref{definition:ruleExpansion} recursively to transform the generated rule
into a set of rules with single (modal) literal in its head.
Afterwards,
similar to the case discussed before,
we can append the \conditionalelement to the body of the rule(s) (\Cref{definition:verificationRuleGeneration}),
as an inference condition,
where appropriate.
Hence,
the statement $ps4$ above can be transformed into the \gls{dl} rules as shown below.

\[
\begin{array}{l@{:~}l}
ps4_1 & \literal{A}(ps4) \defeasible \literal[\OBL]{rel3}[X] \\
ps4_2 & \literal{A}(ps4),\convertviolate{\OBL}{rel3}[X],\violate{ps5} \defeasible\literal[\OBL]{rel4}[X] \\
\end{array}
\]

\subsubsection{Case 2: Referenced Element is a Rule}
Instead, if the referenced element is a rule, then for the case of violation, we have to verify that the rule referenced is either
\begin{enumerate*}[(i)]
\item inapplicable, i.e., there is a literal in its antecedent that is not provable; or
\item the immediate consequent of the rule is defeated or overruled by a conflicting conclusion.
\end{enumerate*} 
While for the case of compliance, 
we  have to verify that the referenced rule is applicable and the immediate consequent of the rule is provable\footnote{In this paper, 
we consider only the case of weak compliance and weak violation, 
and verify only the first (modal) literal that appears in the head of the rule. 
However, 
the method proposed here can be extended easily to support the verification of the cases of strong compliance~\cite{GHM2015} 
and strong violation~\cite{Scannapieco.2011}.}.

\begin{definition}
Let $D=(F,R,\superior)$ be a defeasible theory.
$R^b\subseteq R$ (respectively,
$R^h\subseteq R$) denotes the set of rules
that contains at least one \conditionalelement in their body (head).
\end{definition}
\begin{definition}[Rule Status]
Let $D=(F,R,\superior)$ be a defeasible theory, and let $\Sigma$ be the language of $D$. 
For every $r\in R^b$,
$r_c$ denotes the rule referenced by the \conditionalelement (\xmltag[\lrmlprefix]{Compliance}
or \xmltag[\lrmlprefix]{Violation}).
We define $verifyBody(D)=(F,R',\superior')$ where:

\[
\begin{array}{l}
R'=R \setminus R_b \cup \{
\begin{array}[t]{rl}
r^+_c: & A(r_c)\defeasible \literal{inf}[r_c], \\[.2em]
r^-_c: & \defeasible \literal{\neg inf}[r_c], \\[.2em]
r^-_{cv}:  & \literal{\neg inf}[r_c] \defeasible \literal{violation}[r_c], \\[.6em]
r^+_{cc}: & \literal{inf}[r_c], \comply{C(r_c,1)} \defeasible \literal{compliance}[r_c], \\[.2em]
r^+_{cv}: & \literal{inf}[r_c], \violate{C(r_c,1)} \defeasible \literal{violation}[r_c]~\} \\[.4em]
\end{array} \\
\superior' = \superior \cup \set{r^+_c\superior r^-_c}  

\end{array}
\]

\end{definition}

\noindent For each $r_c$, $\literal{inf}[r_c]$, $\literal{\neg inf}[r_c]$, $\literal{compliance}[r_c]$ 
and $\literal{violation}[r_c]$ are new atoms not in the language of the defeasible theory.
$\literal{inf}[r_c]$ 
and $\literal{\neg inf}[r_c]$ are used to determine whether a rule is \emph{in force} (applicable).
If $r_c$ is in force, we can then verify whether the first literal 
that appears at the head of $r_c$ is compliant 
or violated (represented using the atoms $\literal{compliance}[r_c]$ and $\literal{violation}[r_c]$, 
respectively).

Similar to the case when the referenced object is a literal, depending on where the \conditionalelement is in the rule,
we can append the compliance and violation atoms to the body and head of the rule directly. 
However, unlike the case where the reference element is a literal, 
this time we can append the atoms required directly without any transformation.

\subsection{Implementation}
\label{sec:implementation}

\begin{figure}\centering

{
\def\foldera#1#2{
  \tikzstyle{folderBox} = [fill=white,rectangle,minimum width=1.2cm,minimum height=.7cm,draw,inner sep=0pt];
  \node (#1) [folderBox] at (#2) {};
  \draw [fill=white] ($(#1.north west)+(.5\pgflinewidth,0)$) arc (180:90:.05cm)
    -- ++(.1,0) -- ++ (0,.04) arc (180:90:.05cm) -- ++ (.38cm,0) 
    -- ++ (0,0) arc (90:0:.05cm) -- ++ (0,-.04) 
    -- ++ ($(.5cm,0)+(\pgflinewidth,0)$) 
    -- ++(0,0) arc (90:0:.05cm) -- cycle;
  }

\begin{tikzpicture}[node distance=5em,
nameNode/.style={draw,minimum height=2em,rounded corners=5pt},
labelNode/.style={auto,pos=.5,font=\imagefont[\scriptsize],inner sep=3pt,},
arrLine/.style={Line,->},
]

\node (spindle) [nameNode,minimum width=8em] {\spindle} ;
\node (legalRuleMLx) [nameNode,anchor=north] at ($(spindle.south)+(0,-4em)$) {} ;

\foldera{legalRuleML1}{$(legalRuleMLx)+(7pt,6pt)$}
\foldera{legalRuleML2}{$(legalRuleMLx)+(0pt,0pt)$}
\foldera{legalRuleML3}{$(legalRuleMLx)+(-7pt,-6pt)$}

\node (legalRuleML) [fit=(legalRuleML1)(legalRuleML2)(legalRuleML3)] {};

\node [anchor=west,text width=9.6em,font=\sffamily\scriptsize,inner sep=3pt,xshift=-5pt]
	at (legalRuleML.east) {Defeasible theories\\ represented using different formalisms, such as\\ \spindle DFL, XML,\\ LegalRuleML, etc.
  } ;

\draw [arrLine] ($(legalRuleML.north)+(-5pt,0)$) -- 
	node [labelNode] {Parsing}
	($(spindle.south)+(-5pt,0)$) ;
\draw [arrLine] ($(spindle.south)+(5pt,0)$) -- 
	node [labelNode] {Rendering}
	($(legalRuleML.north)+(5pt,0)$) ;

\end{tikzpicture}
}
\caption{Theory parsing and rendering process}
\label{figure:theoryIOprocess}
\end{figure}
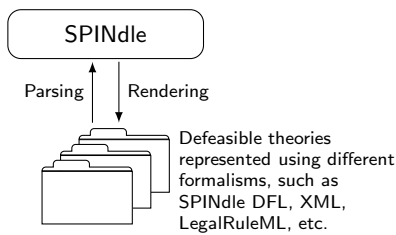

The above transformations can be used to translate legal norms represented using \gls{legalruleml} into \gls{dl} theory 
that we can reason on. We have implemented the above transformations as an extension to the \gls{dl} reasoner \spindle~\cite{Lam.2009}
-- an open-source, Java-vased \gls{dl} reasoner. 
\spindle supports reasoning on both
standard and modal defeasible logic,
such that legal norms represented using \gls{legalruleml} can be parsed into a \spindle \gls{dt} for further processing.
Besides,
we also implemented a theory renderer
so that \glspl{dt} in \spindle can also be exported into \gls{legalruleml} documents
through a rendering process,
as depicted in \Cref{figure:theoryIOprocess}.

To get the 
idea of how the transformations work,
various tests has been carried out to compare the performance of the \gls{legalruleml} theory parser and renderer
with the \spindle DFL theory parser and render~\cite{Lam.2011b},
respectively.
All source code (including \spindle) in the experiments is compiled using the Java SDK 1.8 without any optimization flags.
The times and
memory usage presented in the experiments are those measured by the system functions supported by \gls{jvm},
and was performed on the same lightly loaded Intel Core i5 (3.5GHz) machine operating under macOS 10.13 with 16GB main memory.
Each timing and memory usage datum is the mean of several executions.
There is no substantial variation among the executions,
except as noted.
Time and memory consumed exclude the latency
and overhead caused by \spindle initialization.

The experiments is based on the \gls{tcpc} described in~\cite{Governatori.2016}
which consists of 6 constitutive statements,
78 prescriptive statements
and 10 override statements,
and will be transformed in a defeasible theory with 6 strict rules,
78 defeasible rules
and 10 superiority relations (with 121 literals).
In order to further evaluate the scalability of the \gls{legalruleml} theory parser
and renderer,
we have created a set of synthetic theories by duplicating the set of (all) statements in the original theory
and renamed their keys.

The experiments were carried out as follow.
The \gls{tcpc} theory represented using \gls{legalruleml} will first be parsed 
using the \gls{legalruleml} theory parser into a \spindle \gls{dt}.
Then, 
the generated \gls{dt} will be transformed back into the \gls{legalruleml} formalism to measure the performance of the rendering process.
Note that the \gls{legalruleml} document generated based on the rendering process
will be based only on the information available from the \gls{dt} in \spindle,
which might not be the same as the one that we used as input.
This is due to the fact
that \spindle accepts \gls{dt} represented using different formalisms as input
(as long as a theory parser associated with the formalism is available).
However,
as \gls{legalruleml} is essentially more expressive
and support more features than the \spindle \gls{dt},
some information may be lost
and cannot be captured during the theory generation phase.
Hence,
the theory generated during the rendering process
can only be based on the information available from the \gls{dt},
i.e., the set of rules,
and the penalties/reparations information.
Other details,
such as the metadata of the norms
and information about deontic elements (such as violation and compliance, etc.)
will be lost during the rendering process.

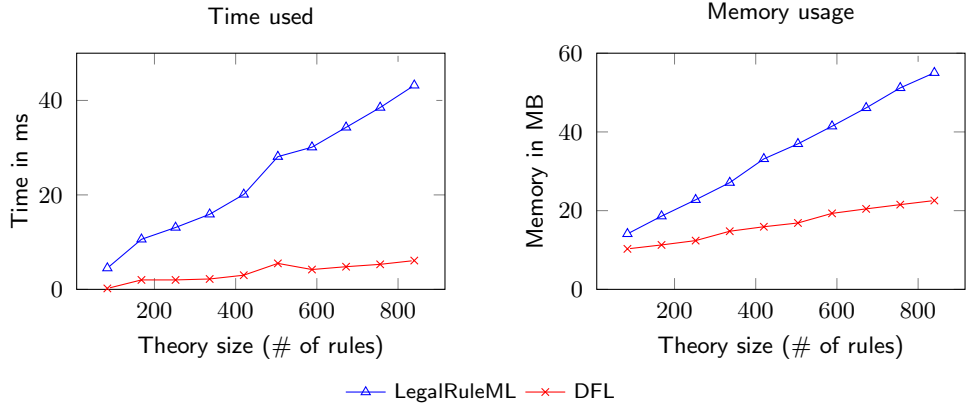
\begin{figure}\centering

{
\pgfplotstableread[col sep=comma]{misc/results_tcpcTheories.csv}\loadedtable
\tikzset{
lrml/.style={mark=triangle,blue},
dfl/.style={mark=x,red},
}
\begin{tikzpicture}
\begin{groupplot}[
width=.48\linewidth,
height=.35\linewidth,
ymin=0,
ymax=60,
xlabel={Theory size (\# of rules)},
x label style={yshift=4pt},
y label style={yshift=-6pt,rotate=0},
group style={group size=2 by 1,horizontal sep=6.2em,},
legend columns=-1,
legend style={draw=none,fill=none,anchor=north,at={(current bounding box.south)}},
legend entries={{\footnotesize LegalRuleML},{\footnotesize DFL}},
legend to name=CombinedLegendAlpha2,
]

\nextgroupplot[title={Time used},ylabel={Time in ms},ymax=50]
\addplot [lrml] table [x=rules,y=lrml_parsing_time] from \loadedtable;\label{lang:lrml}
\addplot [dfl] table [x=rules,y=dfl_parsing_time] from \loadedtable;\label{lang:dfl}

\nextgroupplot[title={Memory usage},ylabel={Memory in MB}]
\addplot [lrml] table [x=rules,y=lrml_parsing_memory] from \loadedtable;
\addplot [dfl] table [x=rules,y=dfl_parsing_memory] from \loadedtable;


\end{groupplot}

\ref{CombinedLegendAlpha2}


\end{tikzpicture}
}
\caption{Performance measurements: Theory parsing}
\label{figure:theoriesParsingResults}
\end{figure}

\begin{figure}\centering

{
\pgfplotstableread[col sep=comma]{misc/results_tcpcTheories.csv}\loadedtable
\tikzset{
lrml/.style={mark=triangle,blue},
dfl/.style={mark=x,red},
}
\begin{tikzpicture}
\begin{groupplot}[
width=.48\linewidth,
height=.35\linewidth,
ymin=0,
ymax=60,
xlabel={Theory size (\# of rules)},
x label style={yshift=4pt},
y label style={yshift=-6pt,rotate=0},
group style={group size=2 by 1,horizontal sep=6.2em,},
legend style={draw=none},
legend columns=-1,
legend style={draw=none,fill=none,anchor=north,at={(current bounding box.south)}},
legend entries={{\footnotesize LegalRuleML},{\footnotesize DFL}},
legend to name=CombinedLegendAlpha2,
]

\nextgroupplot[title={Time used},ylabel={Time in ms},ymax=50]
\addplot [lrml] table [x=rules,y=lrml_rendering_time] from \loadedtable;\label{lang:lrml}
\addplot [dfl] table [x=rules,y=dfl_rendering_time] from \loadedtable;\label{lang:dfl}

\nextgroupplot[title={Memory usage},ylabel={Memory in MB}]
\addplot [lrml] table [x=rules,y=lrml_rendering_memory] from \loadedtable;
\addplot [dfl] table [x=rules,y=dfl_rendering_memory] from \loadedtable;


\end{groupplot}

\ref{CombinedLegendAlpha2}

\end{tikzpicture}
}
\caption{Performance measurements: Theory rendering}
\label{figure:theoriesRenderingResults}
\end{figure}
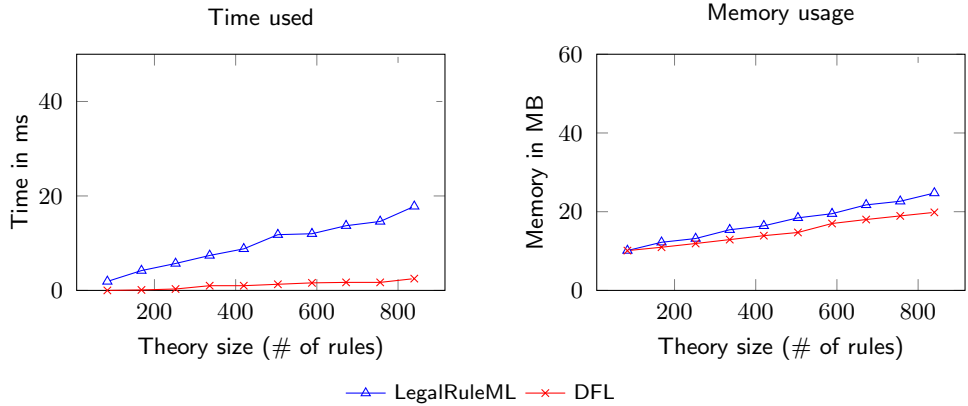

Figures~\ref{figure:theoriesParsingResults}
and~\ref{figure:theoriesRenderingResults} show the performance measured for executing the test theories.
As can be seen from the graphs,
due to its complexity,
it is clear
that the \gls{legalruleml} theory parser
and renderer,
in general,
consume more time and memory
than the \spindle DFL theory parser
and renderer
when parsing
and rendering a \gls{dt}, 
respectively.
The only exception is on the memory consumption
when exporting \glspl{dt} from \spindle
that the two formalisms consume more-or-less the same amount of memory.

This is understandable as \gls{legalruleml} are more expressive
and complex
than the \spindle DFL language,
which requires more time to parse and analyses the internal structure of the document.
As it is generally the case, 
there is always a trade off between expressiveness and efficiency.
Even with the size of the deontic theory 
that we are using,
there is already a performance gap between \gls{legalruleml} 
and the \spindle DFL language 
when parsing the defeasible theory. 
And the same applies, as well, to the rendering process.

In terms of scalability,
except some minor fluctuations 
which may possibly be caused by the operating system
or the \gls{jvm} (such as heap allocation, page swapping, etc),
both theory parser
and renderer perform almost linearly with respect to the size of the theories.

\medskip
As a remark,
the transformation above is conform with the current version of the \gls{legalruleml} specifications~\cite{AthanBGPPW.13}.
However, 
it should be noted that strange results may appear if a \xmltag[\lrmlprefix]{Violation}
(or \xmltag[\lrmlprefix]{Compliance}) element
appears at the head of a statement (i.e., the \xmltag[\rulemlprefix]{then} part of a statement).
For instance,
consider the case where the deontic element \xmltag[\lrmlprefix]{Violation} appears as the only element at the head of a statement.
Then,
it will be transformed into a rule with no head literal,
which is not correct.
In the light of this, 
we believe that additional restriction(s) should be added to the \gls{legalruleml} specification in order to avoid this situation.

\section{Related Work}
\label{section:relatedwork}

The research in the area of e-contracting, business process compliance 
and automated negotiation systems has evolved over the last few years. 
Several new modelling languages have been proposed
and improvements have been made on the existing ones.
On the basis of these modelling languages,
different taxonomies
and semantics of business rules have been developed~\cite{Gordon.2009},
and transformations techniques have emerged facilitating the reasoning process with these languages.

\gls{contractlog}~\cite{Paschke.2005} is a rule-based framework for monitoring 
and execution of service level agreements (SLAs).
It combines rule-based representation of SLAs using Horn clauses and meta programming techniques alternative to contracts defined in natural language or pure programming implementations in programming languages. A rule-based technique called {\em SweetDeal} for representing business contracts that enables the software agent to automatically create, negotiate, evaluate and execute the contract provisions with high degree of modularity is discussed in~\citeN{Grosof.2012}. Their technique builds upon situated courteous logic programs (SCLP) knowledge representation in \gls{ruleml}, and incorporates the process knowledge descriptions whose ontologies are represented in DAML+OIL\footnote{DAM+OIL Reference:~\url{http://www.w3.org/TR/daml+oil-reference/}}. DAML+OIL representations allow handling more complex contracts with behavioral
 provisions that might arise during the execution of contracts. The former has to rely upon multiple formalisms to represent various types of SLA rules e.g. Horn Logic, Event-Calculus, Description Logic\textemdash whereas the latter does not consider normative effects (i.e., the approach is unable to differentiate various types of obligations such as achievement, maintenance and permissions).

\glsfirst{sbvr}~\cite{OMG.2008} is an \gls{omg} standard to represent and fomalise business ontologies, including business rules, facts and business vocabularies. It provides the basis for detailed formal and declarative specifications of business policies and includes deontic operators to represent deontic concepts e.g., obligations, permissions etc. Also, it uses the controlled natural languages to represent legal 
norms~\cite{Gordon.2009}; however, the standard has some shortcomings as the semantics for the deontic notions is underspecified. 
This is because \gls{sbvr} is based on classical first{-}order logic (FOL), 
which is not suitable to represent deontic notions and conflicts 
because FOL has no conceptual relevance to the legal domain~\cite{Herrestad.1991,Hashmi.2017}. 
Also, it cannot handle \gls{ctd} obligations as these cannot be represented by standard deontic logic~(see, \cite{Carmo:02} for details). 

The \glsfirst{lkif}~\cite{ESTRELLAProject.2008}, 
on the other hand, 
is an XML based interchange format language that aims to provide an interchangeable format to represent legal norms in a broad range of application scenarios \textendash~especially in the context of semantic web. 
\gls{lkif} uses XML schemas to represent theories and arguments derived from theories, where a theory in \gls{lkif} is a set of axioms and defeasible inference rules. In addition to these, there are other XML based rule interchange format languages e.g., \gls{swrl}~\cite{Horrocks.2004}, 
RIF~\cite{WCRIFWG.2005}, WSMO~\cite{Roman.2005} and \gls{owl-s}~\cite{OWL.2008} to name but few, see~\cite{Gordon.2009} for more details on the strengths and weaknesses of these languages.

\citeN{Antoniou:2009} presented a model and deontic-based system for representing 
and reasoning policies in multi-agent systems.
Their logical framework extends modal logics with modalities such as \emph{knowledge},
\emph{intention}, \emph{agency} and \emph{obligations},
and, similar to the approach discussed above,
the reasoning part is completed through simulated meta-programming.
Furthermore,
their model also support standard RDF
and RDF schema,
which enable their approach to be compatible with other semantic web technologies.

\citeN{Kontopoulos.2011}, 
on the other hand, 
extended DR-DEVICE~\cite{Bassiliades.2004a} with the capabilities to reason on \gls{mdl} rule bases.
Their approach is based on a \gls{ruleml} like formalism with the support of interactions between modalities,
which may be useful from \gls{semantic web} perspective
where a greedy agent can override its contractual obligations 
and perform what it intends to do.
Similar to our work proposed here,
their approach is based on a series of transformations
that will progressively transform the rule-based into a \gls{dl} theory (for reasoning purpose)
and additional elements,
such as defeaters,
may be added to the theory to resolve issues such as \emph{modality inclusion}.
However,
from legal perspective,
it may not be the same as it is necessary 
that each party complies with the obligations
that are stated in the agreement.
Hence, we have not include this in our framework.
Nevertheless,
as a note,
such features can easily be supported
as it is a built-in feature in \spindle.


\citeN{Steen-2010} proposed an approach
that can automatically transform business rules specifications written using \gls{sbvr} 
into optimised business process models modelled using \gls{bpmn}~\cite{OMG.2008a}
and a domain model represented using UML~\cite{OMG.2000}.
Later,
through utilising a combination of techniques from cognitive linguistics, 
knowledge configuration 
and model-driven engineering,
\citeN{Selway-2015} proposed an approach
that automatically transforms a business specification
into a formal \gls{sbvr} model of vocabulary
and business rules specification.
However, the transformations proposed by~\cite{Steen-2010} are correct by construction but have not been formally verified. In particular, the
transformations related to the process metamodels are generated from the extracted \gls{sbvr} specifications. Besides, even though both approaches enable the generation of suitable formal models from business specifications, the rules transformations and generated models do not include any information on various types of rules as the deontic notions are underspecified in \gls{sbvr}. 

\citeN{Wyner.2013} used C\&C/Boxer~\cite{Bos.2008} to automatically translate normative clauses into semantics representations, 
and compared the results with the logical representation that they created manually (using \gls{dl}),
which narrowed the gap between natural language sources materials
and formal, machine-processable representations.
However, it is not clear how C\&C/Boxer can be used as an abstract representation that is required by C\&C/Boxer.
In contrast,
\citeN{Baget.15} discuss techniques for transforming existential rules into
\gls{datalogp}\textsuperscript{,}\footnote{\gls{datalogp}: a sub-language of \gls{ruleml} \url{http://wiki.ruleml.org/index.php/Rule-Based_Data_Access\#Datalog.2B.2F-}},
\gls{ruleml} and OWL 2 formats.
For the transformation from \gls{datalogp} into \gls{ruleml}, the authors used a fragment of Deliberation \gls{ruleml} 1.01,
which includes positive facts, universally quantified implications, equality,
falsity (and conjunctions) in the heads of implications.
Whereas~\cite{Vojir.2013} transforms the association rules
into Drool Rule Language (DRL) using Lisp-Miner\footnote{Lisp-Miner: \url{http://lispminer.vse.cz}},
\cite{Kamada.2010} proposes a model driven architecture based model to transform \gls{sbvr} compliant business rules extracted
from business contracts of services to compliant executable rules in \gls{fcl}~\cite{Governatori.2005d}.
However,
the former's transformation is limited only to existential rules;
while the latter captures only the business rules (\gls{sbvr} bears only business rules), which may or may not have legal standings. Whilst, \gls{legalruleml} represents legal standings, the \gls{legalruleml}'s temporal notions of enforceability, efficacy and applicability cannot be represented with \gls{sbvr}. In contrast, the approach proposed in this paper enables the translation of defeasible expressions, and various {\em deontic concepts} including the notion of {\em penalty} and {\em chain of reparations}.


\section{Conclusions}
\label{section:conclusions}

In this paper, we have proposed a transformation 
such that (legal) norms represented using \gls{legalruleml} 
can be transformed into \gls{dl} 
which provides us a method for modeling business contracts
and reasoning about them in a declarative way.
Whilst \gls{legalruleml} aims at providing specifications to the legal norms that 
can be represented in a machine readable format,
the major impedance now is the lack of dedicated
and reliable infrastructure that can provide support to such capability.

As a future work,
we are planning to incorporate our technique into some smart-contract enabled systems, such as Ethereum~\cite{Wood.2014}. 
This will extend its language such that,
instead of using programming logics,
users can define their (smart-)contracts in a declarative manner.
Besides, 
we also plan to examine the feasibility of the presented approach
in the context of cyber-physical systems, 
in particular, translation of standards-based regulations for the verification of planned and executed processes for automotive software development.

Since legal documents tend to continuously evolve meaning that new 
rules are added or removed. The addition of new rules into 
legal documents may introduce new rule types with 
varied set of granularity which may increase the complexity of reasoning 
process. Currently, we have applied our transformation approach to 
legal statements 
that appear in very common settings. We plan to continue 
our experiments validating our transforming approach with larger theories 
and rules sets with varied degree of complexity in very concentrated settings.



%

\ignore{\nolinenumbers}

\bibliography{references}

\end{document}